\documentclass[11pt,letterpaper]{article}
\usepackage[T1]{fontenc}
\usepackage[utf8]{inputenc}
\usepackage{lmodern}
\usepackage[margin=0.85in]{geometry}
\usepackage{amsmath,amssymb}
\usepackage{graphicx,booktabs,array}
\usepackage{caption,subcaption}
\usepackage[section]{placeins}
\usepackage{microtype}
\usepackage[numbers,sort&compress]{natbib}
\usepackage[hidelinks]{hyperref}
\title{CFD Correction of Open Tip Clearance Flow in a Compressor Cascade Using VAE Latent Space Adaptation}
\author{\normalsize Xiang Zuo$^1$, Hefang Deng$^2$, Caiyan Chen$^1$, Honglin He$^1$,\\
\normalsize Mingmin Zhu$^1$, Songan Zhang$^{2,*}$, Jinfang Teng$^1$\\[6pt]
\small $^1$School of Aeronautics and Astronautics, Shanghai Jiao Tong University,\\
\small Shanghai 200240, China\\
\small $^2$Global Institute of Future Technology, Shanghai Jiao Tong University,\\
\small Shanghai 200240, China\\[4pt]
\small $^*$Corresponding author: \href{mailto:songanz@sjtu.edu.cn}{songanz@sjtu.edu.cn}}
\date{}
\begin{document}
\maketitle
\begin{abstract}

CFD predictions of open tip clearance flow in compressor cascades are subject to discrepancies relative to experiments, while experimental observations are sparse and high-resolution experimental ground truth is unavailable. This study proposes a non-intrusive correction method based on a variational autoencoder (VAE) and latent-space adaptation. A VAE is first trained using a dataset of 166 parametrically sampled CFD total pressure loss fields to learn a low-dimensional statistical representation of these fields. The VAE is then frozen, and a low-rank latent-space adapter is trained using only 12 paired CFD--experiment operating conditions. An observation operator maps the corrected high-resolution fields to the experimental observation space, allowing supervision to be applied only at the available measurement locations and within the measured pitchwise windows. In the current 12-fold cross-validation, the mean absolute error decreases from 0.1335 to 0.0473, the root mean square error from 0.1717 to 0.0621, and the relative $L_2$ error from 0.5108 to 0.1871. These results indicate that the method improves agreement between CFD predictions and sparse experimental observations of open tip clearance flow without modifying the RANS solver or constructing artificial high-resolution experimental labels.

\end{abstract}

\noindent\textbf{Keywords:} open tip clearance; compressor cascade; CFD correction; variational autoencoder; latent-space adaptation

\section{Introduction}

As aeroengine cores become smaller, blade spans in the rear compressor stages decrease accordingly. However, reductions in absolute tip clearance are generally limited by mechanical design constraints and operating margins, leading to an increase in clearance relative to blade span. A relative clearance exceeding approximately 3.5\% of the blade span is referred to as an open tip clearance \cite{ref1,ref2,ref3}. Compared with conventional small-clearance configurations, open tip clearance flows exhibit stronger interactions among the leakage flow, mainstream, endwall boundary layer, and blade wake. The resulting intense shear and mixing produce a complex total pressure loss distribution at the cascade exit \cite{ref2}. This distribution reflects irreversible dissipation and characterizes the locations, spatial extents, and strengths of different high-loss structures. Accurate measurement or prediction of this distribution is therefore essential for evaluating endwall-region losses and their aerodynamic effects.

The exit total pressure loss distribution in an open tip clearance cascade is strongly dependent on operating conditions. Its location, extent, and magnitude are jointly affected by the clearance size, inlet flow conditions, and endwall motion. The clearance size changes the leakage flow area and leakage mass flow rate. The incidence angle and inlet Mach number affect blade loading and the pressure difference between the pressure and suction surfaces, thereby influencing leakage flow strength and its mixing with the mainstream. The inlet boundary layer and relative motion between the blade and endwall also affect the strength and direction of the leakage flow, as well as the formation, size, and extent of the tip leakage vortex. Qiang et al. \cite{ref4} investigated the effects of a moving endwall on the unsteadiness of tip leakage flow and found changes in its pitchwise transport, vortex morphology, and unsteady behavior. Chen et al. \cite{ref5} further examined the influence of upstream wakes on open tip clearance flow with a moving endwall, identifying changes in exit losses and endwall-region flow structures under different wake conditions. Regarding the inlet boundary layer, Chen et al. \cite{ref6} found that its thickness affects not only the development trajectory of the tip leakage vortex but also the redistribution of losses between the endwall and midspan regions. These studies demonstrate that loss characteristics obtained at a single representative condition cannot adequately describe other geometrical and operating conditions. Exit loss distributions must therefore be obtained over multiple combinations of parameters.

Experimental measurements and numerical simulations are the two principal approaches for obtaining exit total pressure loss distributions under multiple operating conditions. Five-hole probe traverses provide exit total pressure loss measurements, while non-intrusive techniques such as particle image velocimetry can resolve velocities within the clearance, wakes, and local separated-flow structures. Nevertheless, probe size, optical access, traverse time, and experimental cost usually restrict measurements to a limited set of operating conditions and discrete spatial locations \cite{ref3}. In contrast, Reynolds-averaged Navier--Stokes (RANS) methods are relatively inexpensive and well suited to batch simulations across multiple parameters and operating conditions, providing spatially resolved flow-field information. Open tip clearance flow, however, involves interactions among leakage flow, the endwall boundary layer, and blade wakes, including unsteady interactions. RANS-based methods model turbulent fluctuations and the associated mixing processes, introducing uncertainty into local quantitative predictions. Previous experimental--numerical comparisons show that RANS and unsteady RANS (URANS) methods generally reproduce the main trends in tip leakage flow and exit total pressure loss distributions, although local loss magnitudes and spatial distributions may differ from measurements \cite{ref4,ref5,ref6}. These discrepancies depend not only on RANS/URANS modeling approximations but also on the mesh, numerical discretization, and experimental spatial resolution. Moreover, even when RANS predicts overall cascade losses reasonably well, it may misrepresent the relative contributions of individual loss mechanisms, particularly at off-design incidence \cite{ref7}. Higher-fidelity approaches, such as large eddy simulation, resolve unsteady flow structures in greater detail but are too computationally expensive for generating large parametric datasets.

Data-driven methods have recently provided new approaches for integrating numerical simulations and experimental measurements. Existing approaches can be classified into corrections coupled to the numerical solution or governing model equations and non-intrusive corrections applied after the CFD calculation \cite{ref8}. The former typically use data assimilation, field inversion, or machine learning to update flow states, model parameters, or turbulence closures. For example, Ferrero et al. \cite{ref9} used field inversion to obtain RANS model corrections and subsequently developed a data-augmented turbulence model. Such methods can exploit sparse or indirect experimental observations but generally require measurement constraints to be incorporated into the numerical solution or iterative inversion procedure. Non-intrusive methods retain the original solver and instead learn mappings or predictive discrepancies between data sources after the simulations have been completed. Li et al. \cite{ref10} used a multi-fidelity graph neural network to fuse turbomachinery flow fields of different fidelities, demonstrating the potential of data-driven fusion of heterogeneous flow-field information. For open tip clearance flow, however, the high cost of measurements severely limits the number of paired CFD--experiment conditions available for training. A key unresolved issue is how to exploit the spatial structures contained in a larger CFD dataset to establish an effective non-intrusive correction using only a small number of paired experimental cases.

To address this issue, this study proposes a non-intrusive CFD correction method based on a variational autoencoder (VAE) and latent-space adaptation. A VAE is first trained using a dataset of 166 parametrically sampled RANS total pressure loss fields to learn a low-dimensional statistical representation of open tip clearance cascade flows. The encoder and decoder are then frozen, and a lightweight latent-space adapter is trained using only 12 paired CFD--experiment conditions to correct the CFD latent representations. To accommodate differences in resolution and coverage between the experimental measurements and CFD grids, an explicit observation operator maps the corrected high-resolution fields on a regular grid to the actual experimental measurement locations and pitchwise windows. A consistency loss is evaluated in this observation space. The method thus uses sparse measurements to constrain the predicted loss distribution while retaining the spatial resolution and principal structural features of the CFD fields.

\section{Experimental Facility and Data Acquisition}

\begin{figure}[tbp]
\centering
\includegraphics[width=0.95\textwidth]{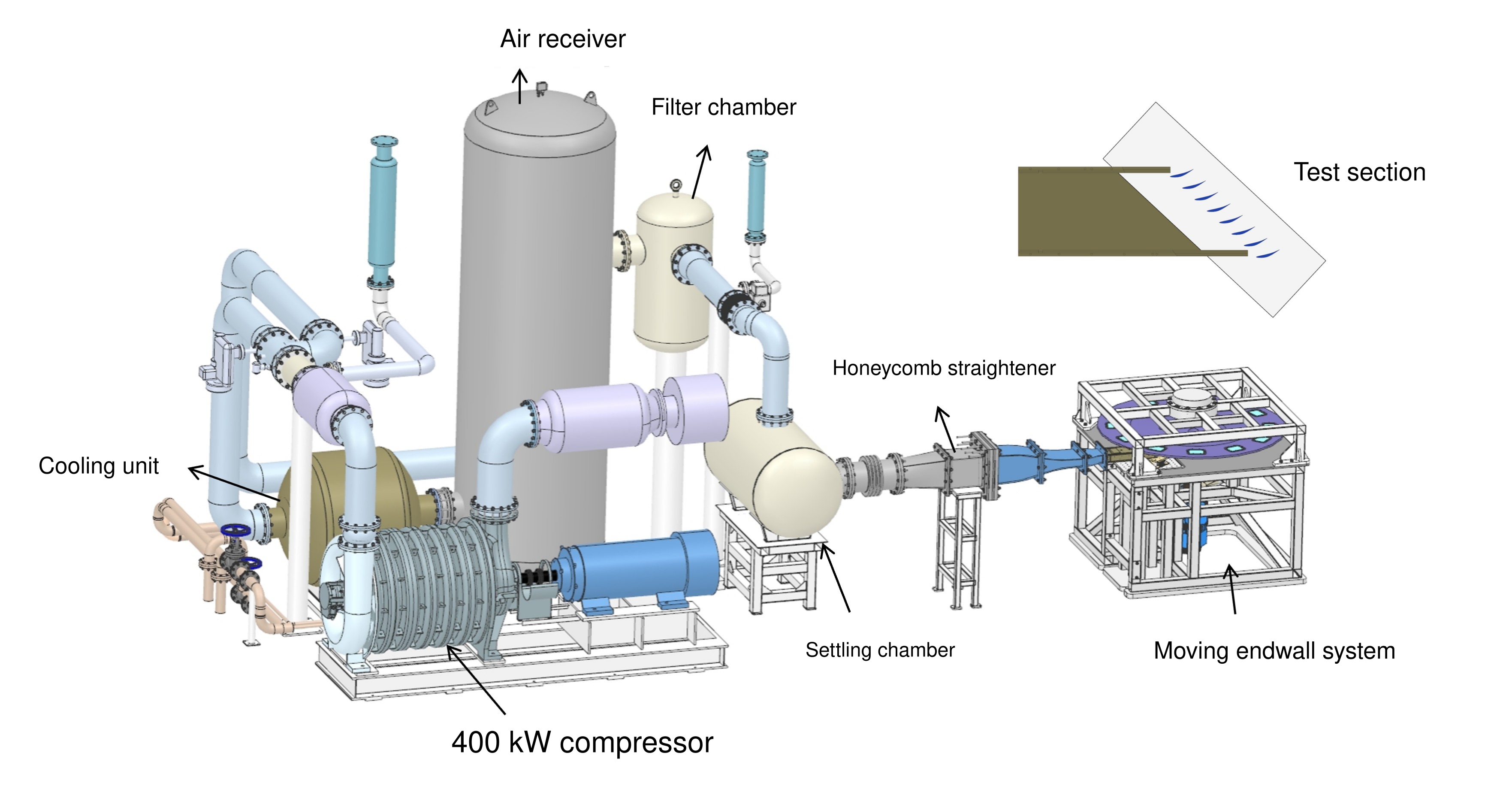}
\caption{Schematic of the experimental facility.}
\label{fig:1}
\end{figure}

The experiments were conducted in the high-speed linear cascade facility with a moving endwall at Shanghai Jiao Tong University. The facility consists primarily of an air supply system, a pressure stabilization and flow conditioning system, a cascade test section, a moving endwall system, a probe traversing system, and a pressure acquisition system. Air is supplied by a centrifugal compressor with a design inlet volumetric flow rate of approximately 180 m\textsuperscript{3}/min. It passes successively through a cooling unit, an air receiver, a filter chamber, a settling chamber, and a honeycomb flow straightener to reduce large-scale disturbances and improve inlet flow uniformity. Different inlet Mach numbers are obtained by adjusting the compressor operating condition. Figure~\ref{fig:1} shows the overall layout of the facility.

\begin{figure}[tbp]
\centering
\includegraphics[width=0.76\textwidth,trim=0 15.873bp 0 0,clip]{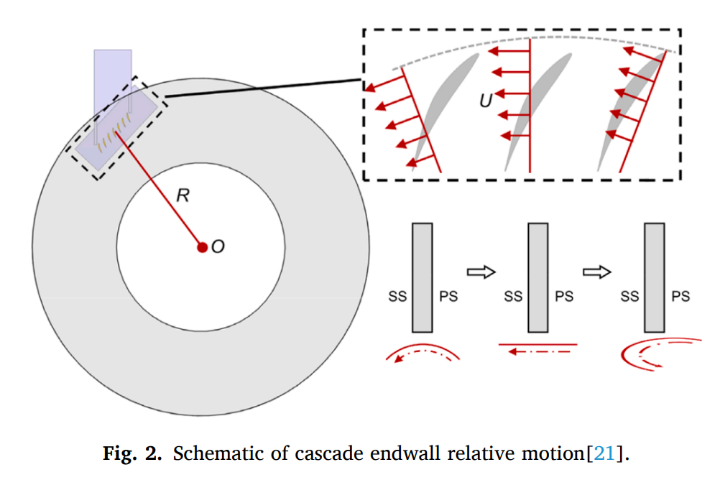}
\caption{Endwall motion in the cascade test section \cite{ref4}.}
\label{fig:2}
\end{figure}

The cascade test section incorporates a large-radius rotating disk whose surface forms the moving endwall adjacent to the blade tips. As shown in Fig.~\ref{fig:2}, the cascade is installed near the disk periphery. Because the disk radius is much larger than the blade chord, its circumferential motion can be approximated locally as translation in the pitchwise direction, thereby simulating the relative motion between compressor stator blades and an adjacent rotating hub \cite{ref4,ref5,ref6}. The cascade is mounted on a vertically adjustable platform, allowing the blade-to-disk clearance to be changed by adjusting the platform height. In the paired dataset used here, the stationary and moving endwall conditions correspond to disk rotational speeds of 0 and 900 rpm, respectively.

\subsection{Test Cascade and Paired Operating Conditions}

\begin{figure}[tbp]
\centering
\includegraphics[width=0.84\textwidth]{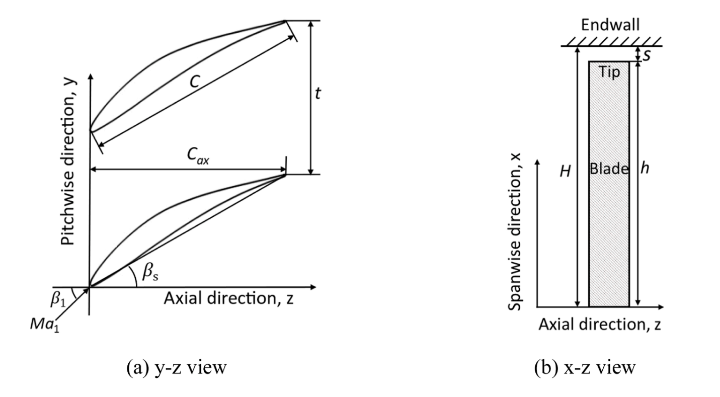}
\caption{CDA blade profile and coordinate definitions.}
\label{fig:3}
\end{figure}

The test configuration is a linear compressor cascade comprising highly loaded controlled-diffusion airfoil (CDA) blades. An open tip clearance is formed between the blade tips and the rotating disk. Table~\ref{tab:1} lists the principal geometrical parameters. Figure~\ref{fig:3} shows the CDA profile and coordinate definitions.

The full experimental program includes measurements of blade surface static pressure, endwall static pressure, the inlet boundary layer, and five-hole probe traverses at the cascade exit. This study uses 12 operating conditions with complete exit traverses and one-to-one correspondence with CFD cases to train and evaluate the CFD correction model.

\begin{table}[tbp]
\centering
\caption{Principal geometrical parameters of the cascade}
\label{tab:1}
\small
\renewcommand{\arraystretch}{1.18}
\begin{tabular}{p{0.43\linewidth}p{0.16\linewidth}p{0.27\linewidth}}
\toprule
Parameter & Symbol & Value \\
\midrule
Blade chord & C & 60 mm \\
Axial chord & $C_{\mathrm{ax}}$ & 51.96 mm \\
Stagger angle & $\beta_s$ & 30\ensuremath{^\circ} \\
Blade pitch & t & 41.27 mm \\
Passage height & H & 70 mm \\
Aspect ratio & h/C & 1.17 \\
\bottomrule
\end{tabular}
\end{table}

The paired operating conditions are defined by the inlet incidence angle, tip clearance, inlet Mach number, and endwall rotational speed, as listed in Table~\ref{tab:2}. They cover incidence angles of 0\ensuremath{^\circ} and 6\ensuremath{^\circ}, tip clearances of 5\% and 7.14\% of blade span, inlet Mach numbers of 0.3 and 0.5, and stationary and moving endwall conditions. Because complete traverses at a clearance of 5\% of blade span were available only at Ma = 0.3, the 12 cases do not constitute a full factorial combination of the four parameters.

\begin{table}[tbp]
\centering
\caption{Paired CFD--experiment operating conditions}
\label{tab:2}
\small
\renewcommand{\arraystretch}{1.18}
\begin{tabular}{ccccc}
\toprule
Case & Incidence (\ensuremath{^\circ}) & Tip clearance (\% span) & Ma & Disk speed (rpm) \\
\midrule
1 & 0 & 5 & 0.3 & 0 \\
2 & 0 & 5 & 0.3 & 900 \\
3 & 0 & 7.14 & 0.3 & 0 \\
4 & 0 & 7.14 & 0.3 & 900 \\
5 & 0 & 7.14 & 0.5 & 0 \\
6 & 0 & 7.14 & 0.5 & 900 \\
7 & 6 & 5 & 0.3 & 0 \\
8 & 6 & 5 & 0.3 & 900 \\
9 & 6 & 7.14 & 0.3 & 0 \\
10 & 6 & 7.14 & 0.3 & 900 \\
11 & 6 & 7.14 & 0.5 & 0 \\
12 & 6 & 7.14 & 0.5 & 900 \\
\bottomrule
\end{tabular}
\end{table}

\subsection{Measurement System and Procedure}

The exit measurement plane is located 0.4C downstream of the blade trailing edge, consistent with the plane used for CFD--experiment comparisons in previous studies \cite{ref5,ref6}. Measurements were obtained using a 2 mm diameter double-L-shaped five-hole probe mounted on a two-axis traversing mechanism with spanwise and pitchwise motion. The probe is inclined upward by 14.2\ensuremath{^\circ} relative to the horizontal plane and installed at an angle of 31.5\ensuremath{^\circ} to the axial direction. At the closest measurement position to the moving endwall, the probe center is 2 mm from the disk surface.

The pitchwise spacing between adjacent measurement locations is 2.5 mm, with 21 locations in total. The spanwise spacing is 2 mm, with 16 locations. The original experimental window therefore contains 16 \ensuremath{\times} 21 = 336 spatial measurement locations for each condition. After the probe reaches a prescribed position, a 3 s settling period allows its position and the pressure signals to stabilize. Data are then acquired continuously for 1 s and time-averaged. The five probe pressures, inlet total pressure, and endwall static pressure are recorded synchronously using an electronic pressure scanning system. The pressure measurement uncertainty is approximately 0.05\% of the sensor full-scale range, and the flow-angle measurement uncertainty is approximately 0.3\ensuremath{^\circ} \cite{ref4}.

\subsection{Experimental Data Processing}

Using the mean inlet total and static pressures as reference quantities, the total pressure loss coefficient at an exit measurement location is defined as

\begin{equation}
\omega = \frac{P_{t,\mathrm{in}}-P_t}{P_{t,\mathrm{in}}-P_{\mathrm{in}}}
\label{eq:1}
\end{equation}

where $P_{t,\mathrm{in}}$ is the inlet total pressure measured by the inlet total-pressure probe, $P_{\mathrm{in}}$ is the inlet endwall static pressure, and $P_t$ is the local total pressure at the exit measurement location. The total pressure loss coefficient quantifies the normalized deficit in local exit total pressure relative to the inlet reference state. It captures the combined losses associated with the blade wake, endwall boundary layer, and tip leakage flow and is the target variable for CFD correction in this study.

The raw five-hole probe measurements form a discrete 16 \ensuremath{\times} 21 observation matrix. Because the experimental traverse does not exactly coincide with the pitchwise periodic region of a single CFD passage, the starting location of the wake is identified from the pitchwise loss distribution before paired samples are constructed. Periodic shifting and cropping are then used to extract a single-pitch experimental window. The resulting window has dimensions 16 \ensuremath{\times} N, where N is 17, 18, or 19 depending on the operating condition. This procedure adjusts only the position and extent of the window; it does not interpolate the measurements into a high-resolution full field. Accordingly, $y_{\mathrm{exp}}$ remains a set of discrete observations at a finite number of locations.

For each paired condition, the data comprise a high-resolution CFD field, the cropped experimental observations, and a condition vector containing the incidence angle, tip clearance, inlet Mach number, and endwall rotational speed. Differences in resolution, spatial coverage, and measurement locations preclude direct pointwise supervision between the two fields. The model therefore uses an explicit observation operator to project the high-resolution predicted field to the experimental measurement locations and evaluates the consistency loss in observation space.

\section{CFD Simulations and Flow Field Dataset Construction}

This section describes the CFD simulations and the processing steps used to construct the flow-field datasets.

\subsection{Computational Domain and Reference Planes}

\begin{figure}[tbp]
\centering
\includegraphics[width=0.85\textwidth]{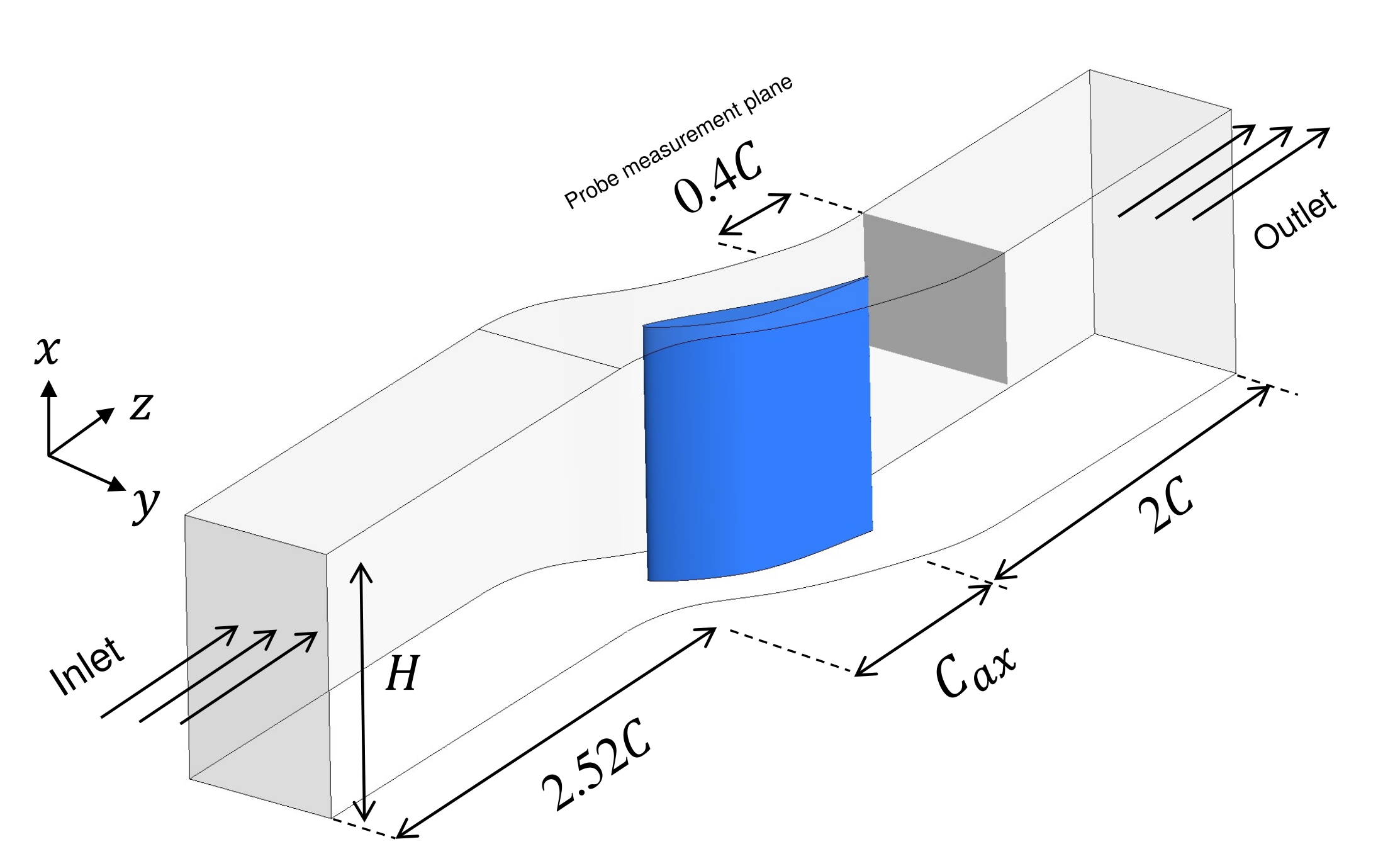}
\caption{Computational domain and reference locations.}
\label{fig:4}
\end{figure}

The CFD model uses the same CDA linear cascade geometry as the experiments. To balance accuracy and the efficiency of large-scale sample generation, a single blade passage is used with translational periodic boundaries in the pitchwise direction. Following the numerical study of the same cascade in Ref. \cite{ref6}, the domain extends from 2.52C upstream of the leading edge to 2C downstream of the trailing edge. The plane 0.45C upstream of the leading edge is used as the inlet aerodynamic reference plane, and the plane 0.4C downstream of the trailing edge is used to extract the total pressure loss field. The latter coincides with the five-hole probe measurement plane and captures the principal flow structures associated with the blade wake, endwall boundary layer, and tip leakage flow. Figure~\ref{fig:4} shows the computational domain and the principal reference locations.

\subsection{Mesh Configuration and Grid Independence}

The 166 parametrically sampled CFD cases used in Stage 1 and the 12 paired CFD--experiment cases used in Stage 2 are computed using the same mesh-generation specification to maintain consistent spatial discretization and numerical resolution. Multiblock structured meshes are generated using NUMECA AutoGrid 5, with local refinement near the blade surfaces, within the endwall boundary layer, and in the tip clearance, where flow gradients are large.

The blade-to-blade region uses the automatic AutoGrid topology with periodic matching. The inlet, blade, and outlet sections contain 29, 89, and 33 streamwise control points, respectively, and 117 grid surfaces are distributed in the spanwise direction. Seventeen mesh layers are placed in the near-wall boundary-layer region around the blade, with additional refinement near the leading edge, trailing edge, and endwalls. The first near-wall cell thickness is 1.0 \ensuremath{\times} 10\textsuperscript{-6} m. A representative mesh contains 1,925,445 nodes, and a multigrid approach is used. Mesh quality checks identify no negative-volume cells. The mesh is regenerated for each tip clearance while retaining the same topology, node allocation, and near-wall refinement principles. This specification is used for all CFD cases in the study.

Qiang et al. \cite{ref4} previously performed a grid-independence study for this CDA cascade using the same structured-mesh framework. Accordingly, a separate grid-independence study is not repeated here.

\subsection{Numerical Method and Boundary Conditions}

The three-dimensional compressible Reynolds-averaged Navier--Stokes equations are solved using ANSYS CFX. All cases use steady RANS. Turbulence closure is provided by the SST k--\ensuremath{\omega} model, coupled with the \ensuremath{\gamma}--$\mathrm{Re}_{\theta}$ transition model to represent boundary-layer transition on the blade surfaces. This model combination has been assessed against blade surface static pressure and exit total pressure loss measurements for the same CDA cascade with a moving endwall and can capture the principal loss structures associated with blade wakes and tip leakage flow \cite{ref5,ref6}.

Total temperature, a total pressure distribution, and flow direction are prescribed at the inlet. The inlet total pressure distribution is based on the measurements in Ref. \cite{ref5}. Different inlet Mach numbers are obtained by matching the inlet and outlet aerodynamic conditions, while incidence is varied by changing the inlet flow angle. An area-averaged static pressure is imposed at the outlet. The blade surfaces and stationary endwalls are treated as adiabatic no-slip walls, and translational periodic conditions are applied to the pitchwise boundaries. The moving endwall is modeled as a planar wall translating in the pitchwise direction. Its speed is calculated from the disk rotational speed and effective radius for the corresponding experimental condition; the wall speed is zero for stationary-endwall cases. Each calculation is terminated after the residuals, inlet--outlet mass balance, and monitored exit flow quantities have stabilized.

\subsection{Parametric Sampling and Generation of the Stage 1 CFD Dataset}

To obtain high-resolution CFD fields covering different aerodynamic states, the inlet incidence angle i, tip clearance s, inlet Mach number Ma, and endwall rotational speed n are selected as the operating-condition parameters. The three continuous variables span i \ensuremath{\in} [\ensuremath{-}7\ensuremath{^\circ}, 7\ensuremath{^\circ}], s \ensuremath{\in} [1.42\%, 7.14\%] of blade span, and Ma \ensuremath{\in} [0.3, 0.8], and are sampled using Latin hypercube sampling. The endwall rotational speed is a discrete two-level variable, n = 0 or 900 rpm, corresponding to stationary and moving endwalls. Compared with a regular full factorial design, Latin hypercube sampling provides more uniform coverage of each continuous parameter dimension for a limited sample count and reduces local clustering. In total, 200 parameter combinations are generated. The cascade geometry and boundary conditions are constructed for each combination, followed by a steady RANS calculation in ANSYS CFX. Table~\ref{tab:3} lists the ranges and types of the parameters.

\begin{table}[tbp]
\centering
\caption{Parameter ranges for the CFD dataset}
\label{tab:3}
\small
\renewcommand{\arraystretch}{1.18}
\begin{tabular}{p{0.27\linewidth}cp{0.32\linewidth}l}
\toprule
Variable & Symbol & Range & Type \\
\midrule
Inlet incidence angle & i & \ensuremath{-}7\ensuremath{^\circ} to 7\ensuremath{^\circ} & Continuous \\
Tip clearance & s & 1.42\% to 7.14\% of blade span & Continuous \\
Inlet Mach number & Ma & 0.3 to 0.8 & Continuous \\
Endwall rotational speed & n & 0 or 900 rpm & Discrete \\
\bottomrule
\end{tabular}
\end{table}

After the simulations, all cases are checked for convergence, data completeness, and valid flow-field output. A total of 166 valid CFD results are retained; the remaining cases are excluded from subsequent model training. The resulting dataset is written as

\begin{equation}
\mathcal{D}_{\mathrm{CFD}}=\{(\mathbf{c}_j,x_{\mathrm{CFD},j})\}
\label{eq:2}
\end{equation}

where $c_j$ = [$i_j$, $n_j$, $\mathrm{Ma}_j$, $s_j$] is the condition vector for case j, containing the incidence angle, rotational speed, inlet Mach number, and tip clearance, in that order, and $x_{\mathrm{CFD},j}$ is the corresponding high-resolution total pressure loss coefficient field at the exit plane. The Stage 1 VAE encodes and reconstructs only $x_{\mathrm{CFD},j}$; the operating-condition parameters are not supplied to the VAE. The condition vector is used only as an auxiliary input to the latent-space adapter in Stage 2.

\subsection{Extraction and Resampling of Total Pressure Loss Fields}

After each valid case has converged, the two-dimensional total pressure loss coefficient field is extracted at 0.4C downstream of the trailing edge. The local coefficient is calculated using Eq.~\eqref{eq:1}, with the mean inlet total and static pressures taken from the reference plane 0.45C upstream of the leading edge. Each CFD case thus provides a high-resolution two-dimensional field containing losses associated with the blade wake, endwall boundary layer, and tip leakage flow.

Although all CFD cases follow the same mesh-generation specification, the original exit-plane data lie on a nonuniform structured grid. Local node positions also vary with the tip clearance because of the geometrical changes. These fields cannot therefore be used directly as uniform neural-network inputs. The CFD plane data are first reordered to establish consistent pitchwise and spanwise coordinate directions. Periodic shifts based on the blade pitch then align the passage and wake locations across cases. Finally, the nonuniform data are interpolated onto a common regular grid of 120 \ensuremath{\times} 68 points. This resolution refers only to the resampled fields supplied to the neural network and should not be confused with the CFD computational mesh size.

\subsection{Datasets for the Two Training Stages}

The two stages use CFD results generated under a common mesh specification but differ in their data composition and training objectives. Stage 1 uses 166 unpaired high-resolution CFD total pressure loss coefficient fields to train the VAE. No experimental field is introduced as a full-field label. Instead, reconstruction loss and KL regularization are used to learn a low-dimensional latent distribution of the CFD fields, enabling the encoder and decoder to compress and reconstruct loss structures across different incidence angles, tip clearances, inlet Mach numbers, and endwall rotational speeds.

Stage 2 uses 12 paired CFD--experiment cases to train the latent-space adapter. Their CFD results are obtained with the same mesh configuration and numerical methods as the Stage 1 cases. Each pair contains a CFD field resampled onto the regular grid at the experimental condition, sparse experimental observations, and the associated condition information. After Stage 1, the VAE encoder and decoder are fixed, and only the adapter is updated. The 12 CFD fields undergo the same coordinate processing and interpolation onto a 120 \ensuremath{\times} 68 regular grid, allowing the frozen encoder to extract latent variables from a consistent data representation. Using these variables and the operating conditions, the adapter predicts corrections such that the decoded high-resolution field agrees with the measured total pressure loss coefficients at the experimental measurement locations.

The latent-space correction learned in Stage 2 represents the aggregate discrepancy between paired CFD predictions and experimental observations. This discrepancy includes the combined effects of turbulence-model approximations, idealized boundary conditions, numerical discretization, and experimental measurement. The method is therefore described as an observation-based correction of the overall CFD discrepancy, rather than a correction of a single error source or a particular turbulence-model parameter.

\section{CFD Correction Using a VAE and Latent Space Adaptation}

The problem exhibits a pronounced asymmetry between the available data sources. CFD provides spatially complete, high-resolution fields at relatively low cost and is therefore suitable for learning spatial structures such as wakes, endwall boundary layers, and tip leakage loss regions. However, turbulence modeling, idealized boundary conditions, and numerical discretization generally introduce systematic discrepancies relative to experiments. Five-hole probe measurements provide more direct physical evidence but are available only at a limited number of locations and within restricted pitchwise windows, rather than as full-field labels at the CFD resolution. Directly training a high-resolution predictor on these sparse measurements would therefore lack complete supervision and could produce poorly supported predictions in unobserved regions.

\begin{figure}[tbp]
\centering
\begin{subfigure}{0.80\textwidth}
\centering
\includegraphics[width=\linewidth,height=0.34\textheight,keepaspectratio]{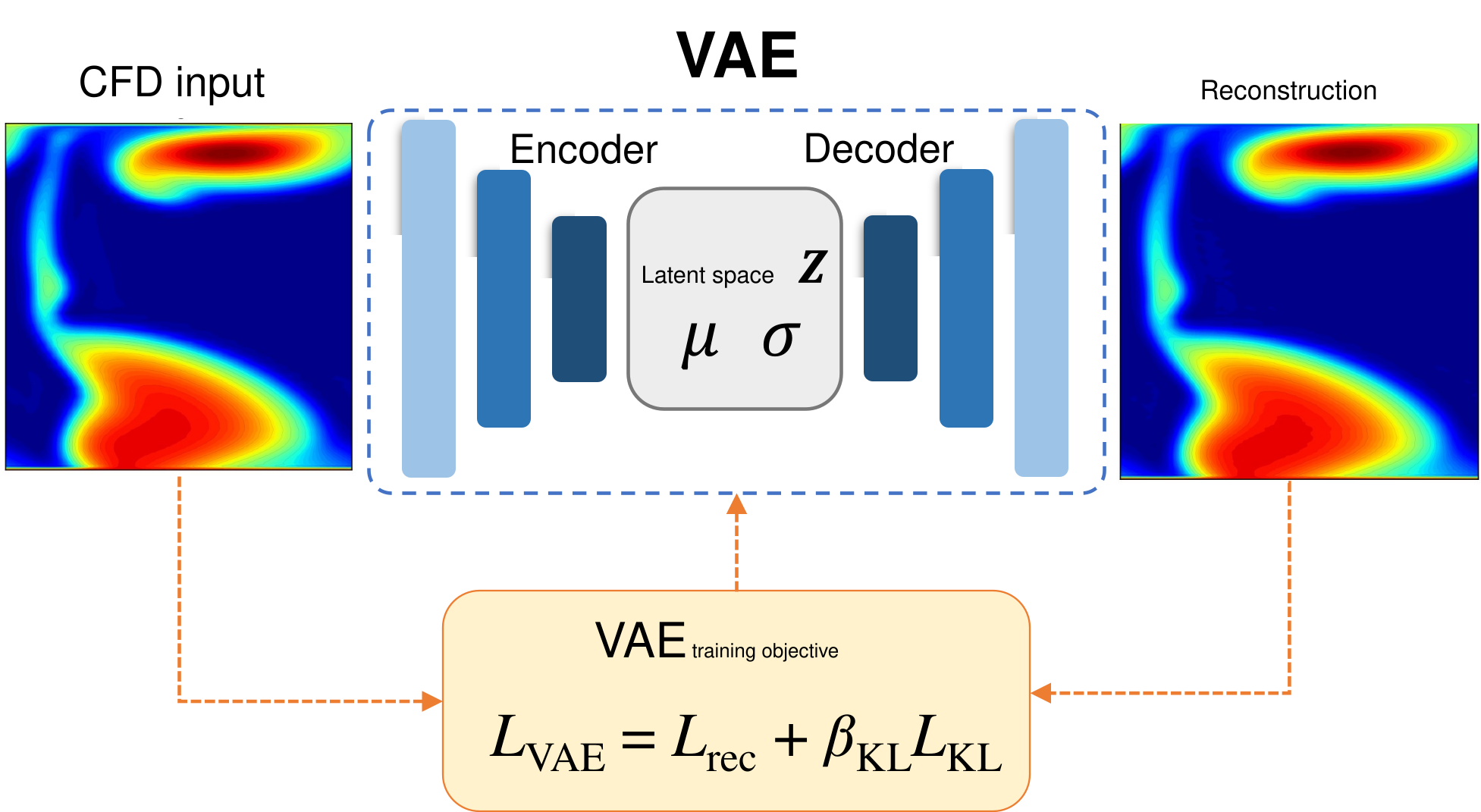}
\caption{Stage 1 architecture and training procedure}
\end{subfigure}
\par\medskip
\begin{subfigure}{0.80\textwidth}
\centering
\includegraphics[width=\linewidth,height=0.34\textheight,keepaspectratio]{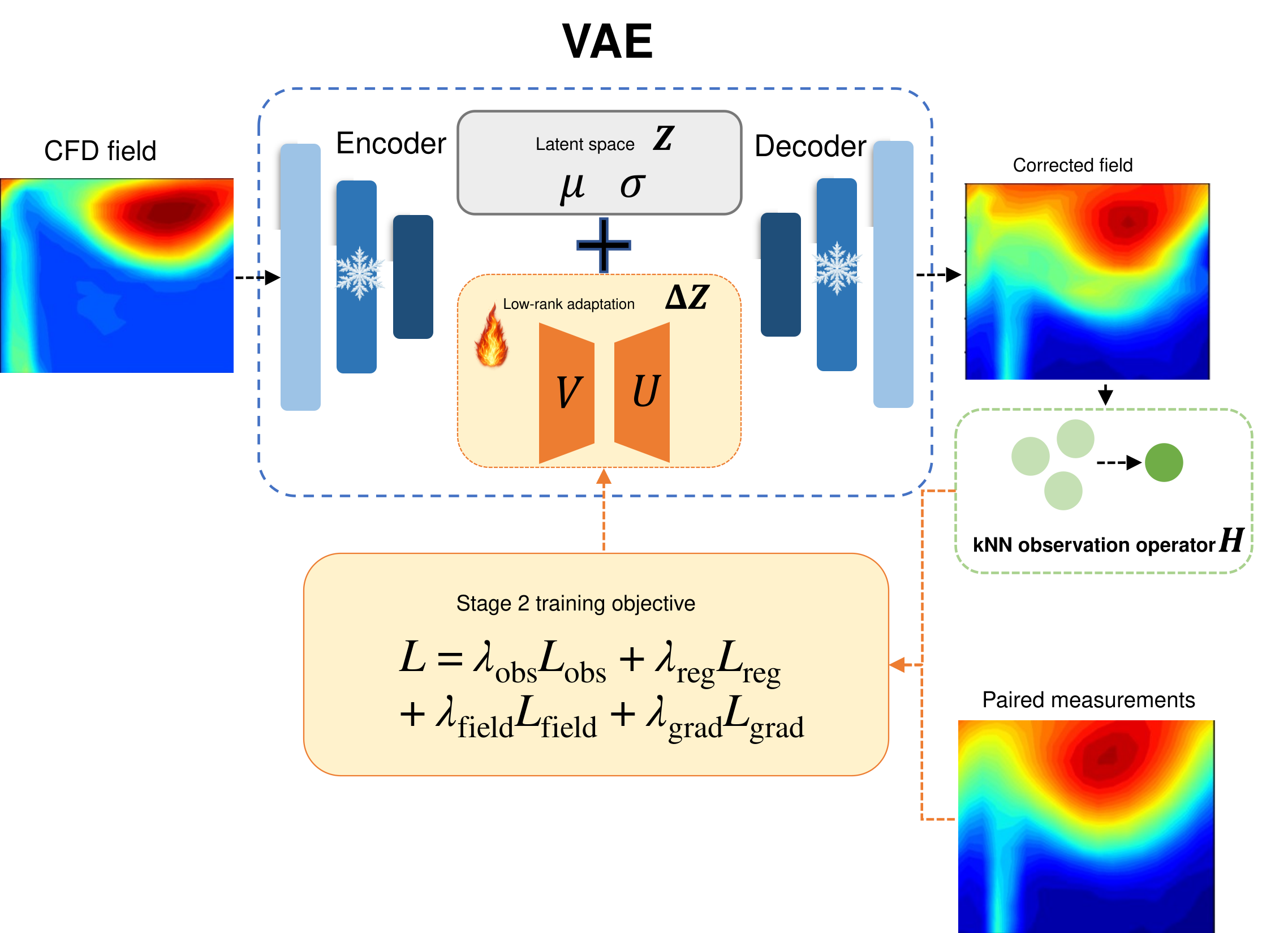}
\caption{Stage 2 architecture and training procedure}
\end{subfigure}
\par\medskip
\caption{Architecture of the two-stage CFD correction model.}
\label{fig:5}
\end{figure}

To address this asymmetry, a two-stage model is constructed: learning the statistical structure of CFD fields, followed by correction using experimental observations. In Stage 1, a VAE \cite{ref11} is trained on the larger CFD dataset to learn a low-dimensional statistical representation of high-resolution loss fields. In Stage 2, the trained encoder and decoder are frozen, and a lightweight latent-space adapter is trained using the small paired CFD--experiment dataset. This approach retains the complete spatial structures provided by CFD while focusing the limited experimental supervision on the aggregate CFD--experiment discrepancy. It is a non-intrusive black-box correction applied after CFD and does not modify the RANS equations, boundary conditions, or turbulence model. Figure~\ref{fig:5} illustrates the model and its two-stage training procedure.

For a CFD field $x_{\mathrm{cfd}}$ resampled onto the regular grid, the encoder first obtains its latent representation $z_{\mathrm{cfd}}$. The adapter then predicts a latent correction $\Delta z$ conditioned on c = [i, n, Ma, s], and the frozen decoder produces the corrected field:

\begin{equation}
\begin{aligned}z_{\mathrm{cfd}}&=E(x_{\mathrm{cfd}}), &\Delta z&=A(z_{\mathrm{cfd}},\mathbf{c}),\\z_{\mathrm{corr}}&=z_{\mathrm{cfd}}+\Delta z, &x_{\mathrm{corr}}&=D(z_{\mathrm{corr}}).\end{aligned}
\label{eq:3}
\end{equation}

Here, E, D, and A denote the encoder, decoder, and adapter, respectively; c contains the incidence angle, rotational speed, inlet Mach number, and tip clearance. Rather than changing individual grid-point values directly, latent-space adaptation adjusts the global field representation learned by the VAE, which helps preserve spatial continuity in wakes and high-loss regions. The output $x_{\mathrm{corr}}$ retains the spatial resolution of the resampled CFD field.

\subsection{A VAE Based Generative Prior for CFD Fields}

Stage 1 aims to extract a transferable prior for flow-field structure from the 166 parametrically sampled CFD cases. After coordinate reordering, periodic shifting, and interpolation onto a regular grid, each sample is represented as a single-passage total pressure loss coefficient field on the common grid. The encoder uses multiscale convolutional residual blocks to extract loss-distribution features at different spatial scales, while the decoder reconstructs the original field from the latent variables. Unlike a conventional autoencoder, a VAE learns a latent probability distribution conditioned on the input field rather than mapping each sample to an isolated deterministic code. This promotes a continuous latent space and provides a smooth, constrained representation for subsequent correction with limited paired data.

The encoder outputs the mean \ensuremath{\mu} and log variance of a latent Gaussian distribution. During training, latent samples are obtained using the reparameterization trick:

\begin{equation}
z=\mu+\sigma\odot\varepsilon,\qquad\varepsilon\sim\mathcal{N}(0,I)
\label{eq:4}
\end{equation}

where \ensuremath{\varepsilon} follows a standard normal distribution. Stochastic sampling during training encourages neighboring flow fields to have continuous latent representations. The distribution mean is used for validation and Stage 2 correction to avoid additional variability from random sampling. The latent variables retain the spatial organization needed to recover local structures such as wakes and high-loss regions, while substantially reducing the degrees of freedom of the subsequent correction problem.

The VAE training objective combines a field reconstruction error with Kullback--Leibler (KL) divergence regularization:

\begin{equation}
\mathcal{L}_{\mathrm{VAE}}=\mathcal{L}_{\mathrm{rec}}+\beta_{\mathrm{KL}}\mathcal{L}_{\mathrm{KL}},\qquad\mathcal{L}_{\mathrm{rec}}=\operatorname{mean}\!\left[(x_{\mathrm{cfd}}-D(z))^2\right]
\label{eq:5}
\end{equation}

Here, $\mathcal{L}_{\mathrm{rec}}$ measures the difference between the reconstructed field and the input CFD field, encouraging the latent variables to preserve the dominant flow structures. $\mathcal{L}_{\mathrm{KL}}$ constrains the latent distribution toward a standard normal distribution and discourages isolated sample encodings. Together, these terms balance reconstruction accuracy against latent-space regularity. Experimental data are not used in this stage; the learned representation is therefore a generative prior for CFD fields, rather than a fit to experimental results. Table~\ref{tab:4} summarizes the Stage 1 training settings.

\begin{table}[tbp]
\centering
\caption{Stage 1 VAE training settings}
\label{tab:4}
\small
\renewcommand{\arraystretch}{1.18}
\begin{tabular}{p{0.32\linewidth}p{0.57\linewidth}}
\toprule
Item & Setting \\
\midrule
Dataset split & 141 training cases; 25 validation cases \\
Optimizer & AdamW \\
Initial learning rate & 1 \ensuremath{\times} 10\textsuperscript{-4} \\
Batch size & 8 \\
Maximum epochs & 300 \\
$\beta_{\mathrm{KL}}$ & 1 \ensuremath{\times} 10\textsuperscript{-4} \\
\bottomrule
\end{tabular}
\end{table}

\subsection{Low Rank Latent Space Adapter}

Only 12 paired CFD--experiment cases are available in Stage 2. Fine-tuning the entire VAE would re-estimate many network parameters from very few samples, potentially disrupting the learned CFD prior and causing overfitting or nonphysical perturbations in unobserved regions. The encoder and decoder are therefore frozen, and only a compact latent-space adapter is trained. Experimental information is thereby restricted to a residual correction of the original CFD latent representation.

The adapter takes $z_{\mathrm{cfd}}$ and c as inputs and predicts $\Delta z$. The condition vector distinguishes the effects of incidence, clearance, Mach number, and endwall rotational speed on the CFD--experiment discrepancy. The residual formulation anchors the correction to the original CFD representation, encouraging the model to learn only changes supported by observations rather than generate an arbitrary new field.

To further limit the trainable degrees of freedom, the adapter uses a low-rank mapping. It first projects the latent representation into a lower-dimensional subspace, nonlinearly combines it with the operating-condition information, and then maps the result back to the original latent space. The nonlinear mapping uses SiLU activations and dropout regularization \cite{ref12,ref13}. The low-rank constraint concentrates the correction along a small number of dominant directions. Together with the frozen VAE, it restricts the solution space while balancing experimental agreement and structural stability of the corrected field.

\subsection{Observation Operator for Sparse Experimental Data}

\begin{figure}[tbp]
\centering
\includegraphics[width=0.91\textwidth]{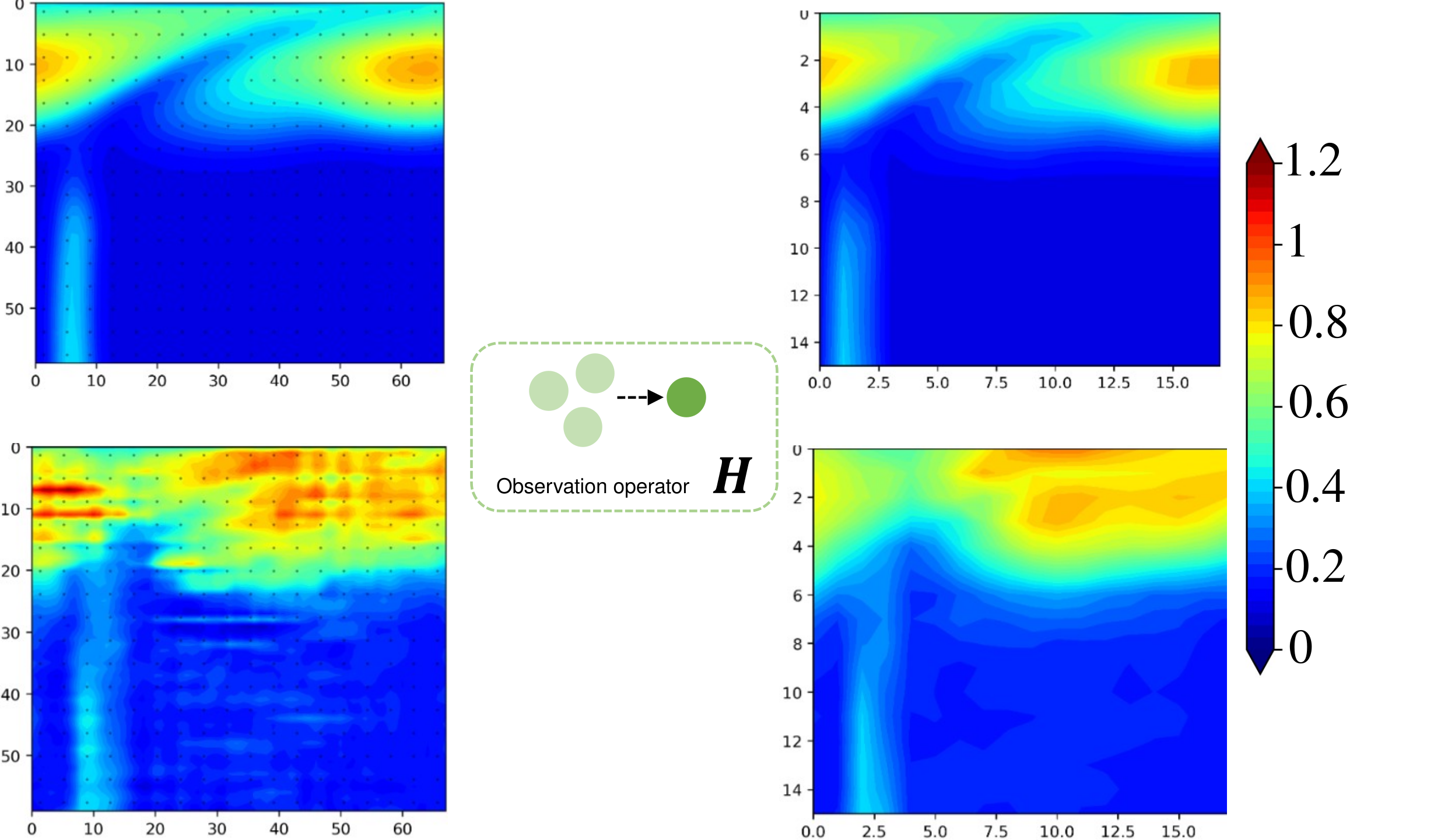}
\caption{Schematic of the observation operator.}
\label{fig:6}
\end{figure}

Because experimental data cover only sparse locations or cropped pitchwise windows, no artificial high-resolution experimental labels are constructed, and $x_{\mathrm{corr}}$ is not compared with measurements across the entire field. Instead, an observation operator H projects the corrected high-resolution field onto the actual measurement locations and observation window. Supervision is thus based exclusively on experimentally observable information. Figure~\ref{fig:6} illustrates the mapping between the regular CFD field and the experimental window.

The mapping first aligns the coordinate origin, spatial directions, and pitchwise periodic position of the CFD field and experimental window. It then applies inverse-distance weighting using the k nearest neighbors \cite{ref14}. For an experimental location $p_m$, a neighboring set $\mathcal{N}_m$ is selected from the regular CFD grid, and interpolation weights are determined from the distances:

\begin{equation}
H(x)_m=\sum_{j\in\mathcal{N}_m}w_{mj}x_j,\qquad w_{mj}=\frac{(d_{mj}+\varepsilon)^{-1}}{\sum_{l\in\mathcal{N}_m}(d_{ml}+\varepsilon)^{-1}}
\label{eq:6}
\end{equation}

The number of neighbors is k = 4. The neighborhood connectivity and interpolation weights are determined in advance from the spatial coordinates and remain fixed during training. The mapping is nevertheless differentiable with respect to the predicted field values, so measurement errors can be backpropagated through the decoder to the adapter. The model thus learns a high-resolution field constrained by actual measurements, without artificially augmenting the sparse experimental observations.

\subsection{Joint Loss Function and Structural Constraints}

Minimizing only the measurement error could lead to large latent shifts that improve local agreement while propagating unconstrained changes into unobserved regions. Stage 2 therefore constrains the latent correction magnitude, full-field deviation, and local structural continuity in addition to experimental consistency. Defining $x_{\mathrm{recon}}$ = D($z_{\mathrm{cfd}}$) and e = H($x_{\mathrm{corr}}$) \ensuremath{-} $y_{\mathrm{exp}}$, the total loss is

\begin{equation}
\mathcal{L}=\lambda_{\mathrm{obs}}\mathcal{L}_{\mathrm{obs}}+\lambda_{\mathrm{reg}}\mathcal{L}_{\mathrm{reg}}+\lambda_{\mathrm{field}}\mathcal{L}_{\mathrm{field}}+\lambda_{\mathrm{grad}}\mathcal{L}_{\mathrm{grad}}
\label{eq:7}
\end{equation}

The observation term $\mathcal{L}_{\mathrm{obs}}$ uses an $L_1$ loss to measure discrepancies at the actual measurement locations and provides the principal supervision for CFD correction. $\mathcal{L}_{\mathrm{reg}}$ limits the magnitude of $\Delta z$ to prevent excessive departure from the CFD latent representation. $\mathcal{L}_{\mathrm{field}}$ constrains the overall difference between $x_{\mathrm{corr}}$ and the VAE reconstruction $x_{\mathrm{recon}}$, primarily to stabilize unobserved regions. $\mathcal{L}_{\mathrm{grad}}$ constrains first-order gradients of the error field to suppress local jaggedness and high-frequency oscillations. These regularizers do not require the corrected field to reproduce the original CFD field; instead, they provide stability where experimental information is insufficient. Table~\ref{tab:5} lists the loss weights and their roles.

\begin{table}[tbp]
\centering
\caption{Stage 2 loss weights}
\label{tab:5}
\small
\renewcommand{\arraystretch}{1.18}
\begin{tabular}{p{0.33\linewidth}cp{0.44\linewidth}}
\toprule
Loss term & Weight & Role \\
\midrule
Observation consistency $\mathcal{L}_{\mathrm{obs}}$ & 1.0 & Fit the experimental measurements \\
Latent regularization $\mathcal{L}_{\mathrm{reg}}$ & 1 \ensuremath{\times} 10\textsuperscript{-4} & Limit the latent correction magnitude \\
Full-field preservation $\mathcal{L}_{\mathrm{field}}$ & 1 \ensuremath{\times} 10\textsuperscript{-4} & Stabilize unobserved regions \\
Gradient constraint $\mathcal{L}_{\mathrm{grad}}$ & 0.005 & Preserve local first-order continuity \\
\bottomrule
\end{tabular}
\end{table}

\subsection{Leave One Condition Out Training with Limited Paired Data}

With only 12 paired operating conditions, a fixed train--test split would make the evaluation highly sensitive to individual cases. A 12-fold leave-one-condition-out scheme is therefore used: the adapter is trained on 11 conditions, and transfer performance is evaluated on the remaining condition, which is excluded from parameter updates. The procedure is repeated until each condition has been held out once. The adapter is reinitialized in every fold, while the encoder and decoder remain frozen throughout. This setup is intended to assess transfer to unseen paired conditions rather than agreement with the training data. Parameters are updated using the AdamW optimizer \cite{ref15}.

To ensure independent evaluation, checkpoint selection and early stopping should, in principle, rely exclusively on information from the training portion, with the held-out condition evaluated only once after training. The current results are retained as a record of methodological feasibility; the final quantitative conclusions will be updated after the model-selection procedure has been strictly fixed.

\subsection{Evaluation Metrics and Visualization}

Because experimental ground truth is available only in observation space, quantitative evaluation follows the same spatial definition as training. The uncorrected and corrected predictions are H($x_{\mathrm{recon}}$) and H($x_{\mathrm{corr}}$), respectively, and are compared with $y_{\mathrm{exp}}$ using mean absolute error (MAE), root mean square error (RMSE), and relative $L_2$ error. MAE measures the average absolute discrepancy at the measurement locations, RMSE is more sensitive to large errors, and relative $L_2$ error measures the error relative to the overall magnitude of the experimental field. The improvement rate is defined as

\begin{equation}
R_{\mathrm{imp}}=\frac{M_{\mathrm{before}}-M_{\mathrm{after}}}{M_{\mathrm{before}}}\times100\%
\label{eq:8}
\end{equation}

Visualizations first compare the experimental observations, predictions before and after correction, and corresponding absolute errors to assess improvement within the observable region. The complete high-resolution corrected field is also displayed to inspect unobserved regions for conspicuous anomalous changes. This full field should be interpreted as a CFD post-processing result constrained by the CFD statistical representation and experimental observations, rather than as high-resolution experimental ground truth that can be validated point by point.

\section{Results and Discussion}

This section first evaluates the Stage 1 VAE reconstruction of the parametrically sampled CFD total pressure loss fields to determine whether the latent representation preserves the dominant loss structures. The 12-fold results are then used to quantify the adapter's correction performance for unseen paired conditions. Finally, representative cases are examined by comparing measurements with the predictions before and after correction, focusing on changes in local high-loss structures and the remaining discrepancies.

\subsection{Stage 1 Reconstruction Performance}

\begin{figure}[tbp]
\centering
\includegraphics[width=0.70\textwidth]{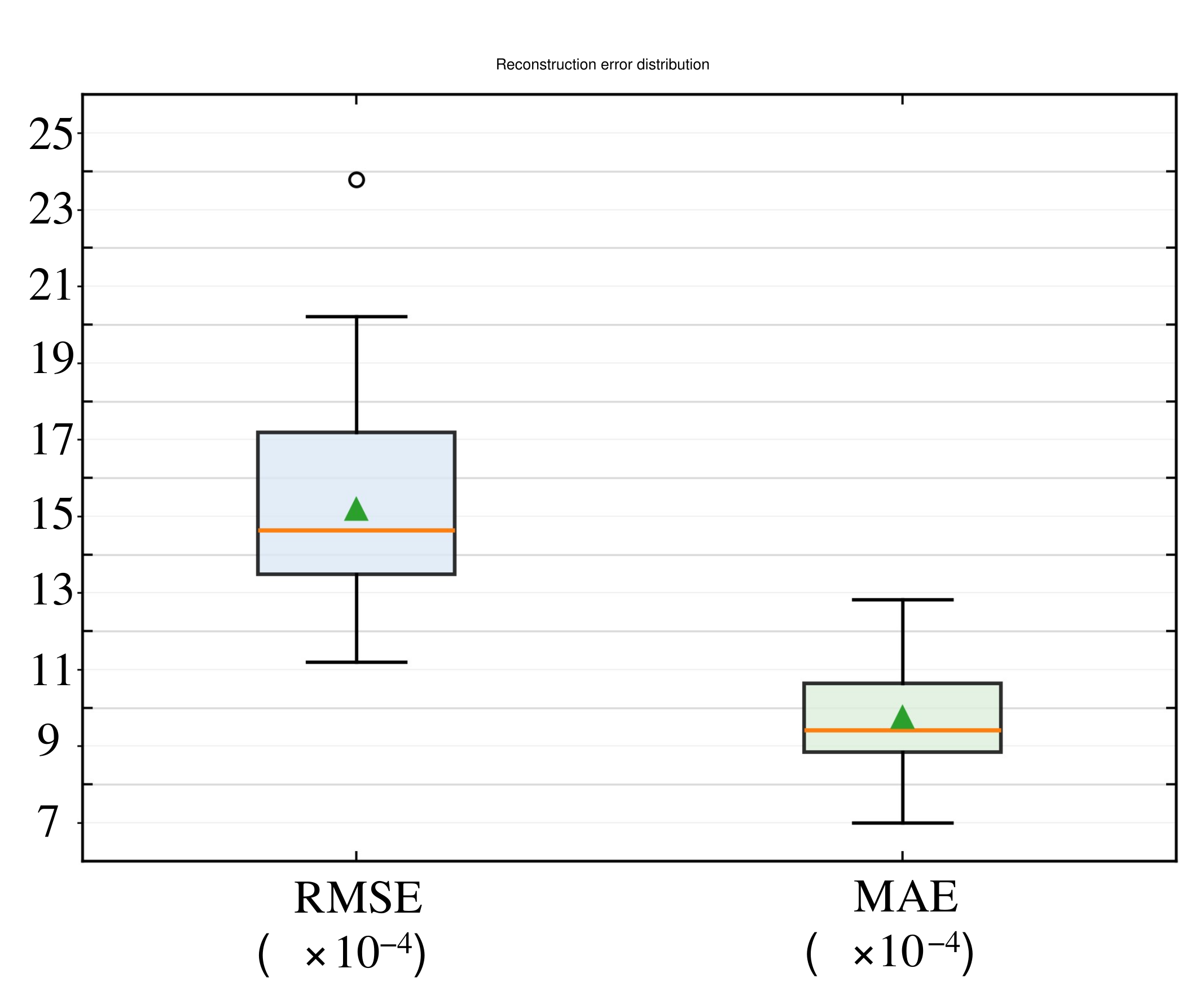}
\caption{Distribution of Stage 1 reconstruction errors on the 25 held-out CFD samples.}
\label{fig:7}
\end{figure}

Stage 1 reconstruction performance is evaluated using 25 held-out CFD samples that are not used for model parameter updates. Figure~\ref{fig:7} presents box plots of the RMSE and MAE for these samples.

\begin{figure}[tbp]
\centering
\includegraphics[width=1.0\textwidth]{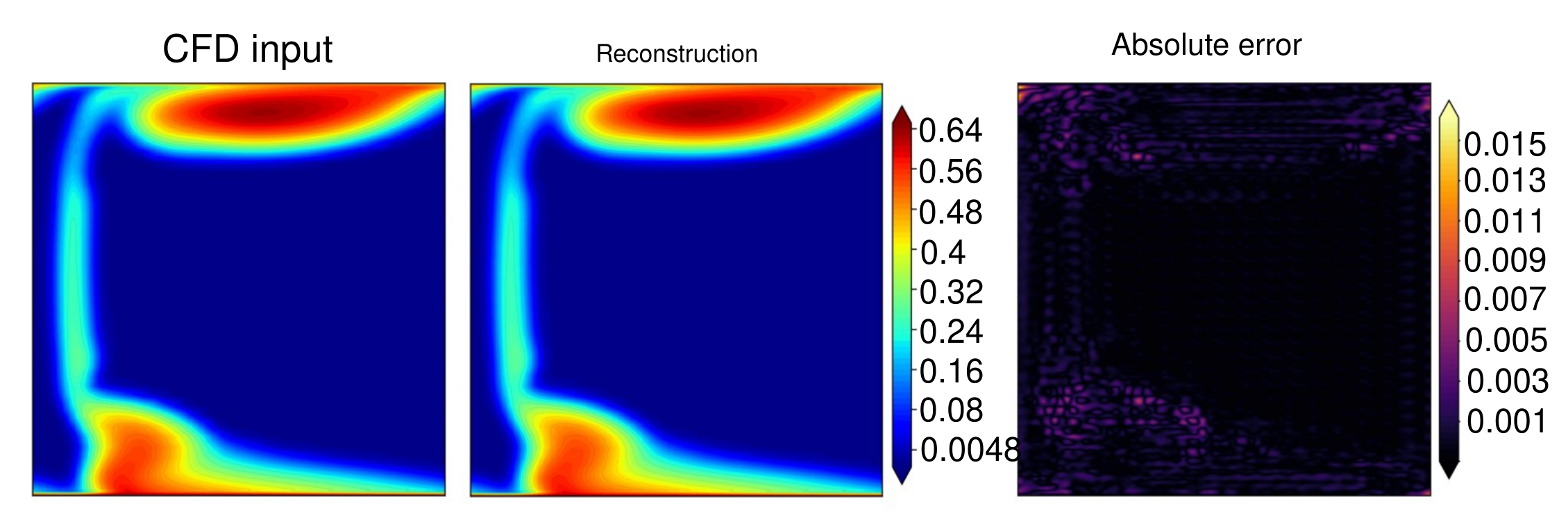}
\caption{Representative Stage 1 reconstruction.}
\label{fig:8}
\end{figure}

Figure~\ref{fig:8} further compares the original CFD field, the VAE reconstruction, and the absolute error for a representative sample. The VAE preserves the principal locations and extents of high-loss regions near the endwall, as well as their overall spatial gradients. Reconstruction errors are concentrated mainly near loss peaks and the edges of high-loss regions. These results indicate that the Stage 1 latent space represents the dominant variations in the CFD fields, providing a basis for Stage 2 latent-space correction with the decoder frozen.

\subsection{Stage 2 Correction Performance}

Table~\ref{tab:6} summarizes the 12-fold results. The values following \ensuremath{\pm} are population standard deviations across the 12 folds and describe variability among operating conditions.

The mean and standard deviation of the foldwise MAE improvement rate are 63.00\% and 12.77\%, respectively; the corresponding values for RMSE are 62.46\% and 12.69\%. Both MAE and RMSE improve for all 12 conditions, indicating reduced observation-space errors in every held-out case under the current checkpoint-selection procedure. The improvement varies considerably across folds, with the lowest values around 34\%, demonstrating a substantial dependence on operating conditions.

\begin{table}[tbp]
\centering
\caption{Summary of the current 12-fold leave-one-condition-out results}
\label{tab:6}
\small
\renewcommand{\arraystretch}{1.18}
\begin{tabular}{lccc}
\toprule
Metric & \shortstack{Before correction\\(mean \ensuremath{\pm} SD)} & \shortstack{After correction\\(mean \ensuremath{\pm} SD)} & \shortstack{Improvement (\%)\\(mean \ensuremath{\pm} SD)} \\
\midrule
MAE & 0.133478 \ensuremath{\pm} 0.022167 & 0.047329 \ensuremath{\pm} 0.012506 & 63.00 \ensuremath{\pm} 12.77 \\
RMSE & 0.171729 \ensuremath{\pm} 0.023534 & 0.062099 \ensuremath{\pm} 0.014686 & 62.46 \ensuremath{\pm} 12.69 \\
Relative $L_2$ & 0.510767 \ensuremath{\pm} 0.051110 & 0.187061 \ensuremath{\pm} 0.050893 & --- \\
\bottomrule
\end{tabular}
\end{table}

\subsection{Corrections in the Experimental Window and Full Field}

\begin{figure}[tbp]
\centering
\begin{subfigure}{0.88\textwidth}
\centering
\includegraphics[width=\linewidth,height=0.19\textheight,keepaspectratio]{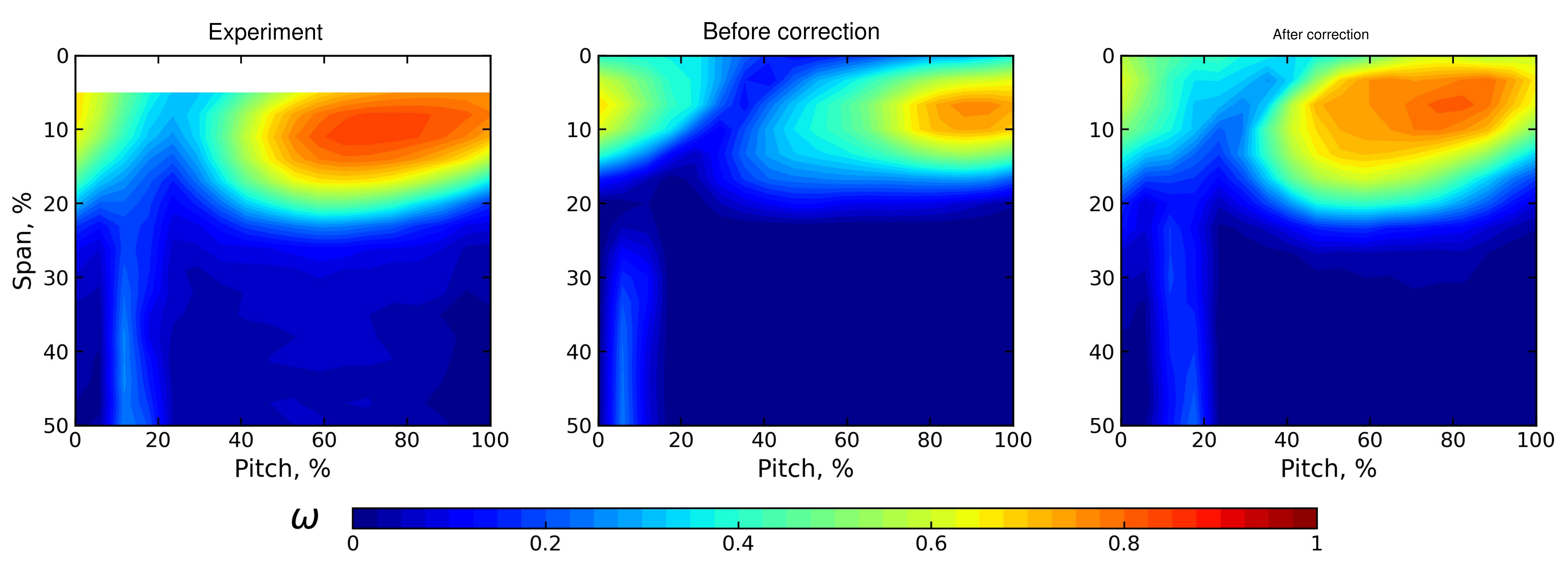}
\caption{i = 0\ensuremath{^\circ}, s = 5\% of blade span, Ma = 0.3, n = 0 rpm}
\end{subfigure}
\par\smallskip
\begin{subfigure}{0.88\textwidth}
\centering
\includegraphics[width=\linewidth,height=0.19\textheight,keepaspectratio]{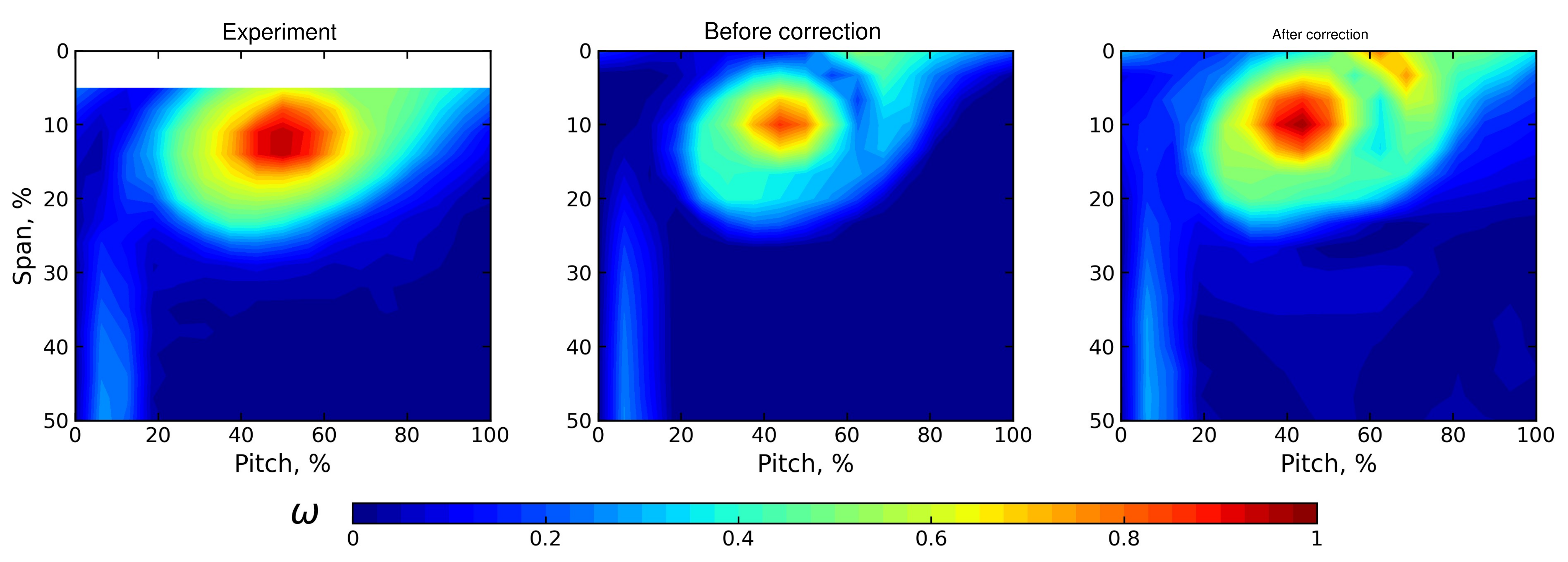}
\caption{i = 0\ensuremath{^\circ}, s = 7.14\% of blade span, Ma = 0.5, n = 900 rpm}
\end{subfigure}
\par\smallskip
\begin{subfigure}{0.88\textwidth}
\centering
\includegraphics[width=\linewidth,height=0.19\textheight,keepaspectratio]{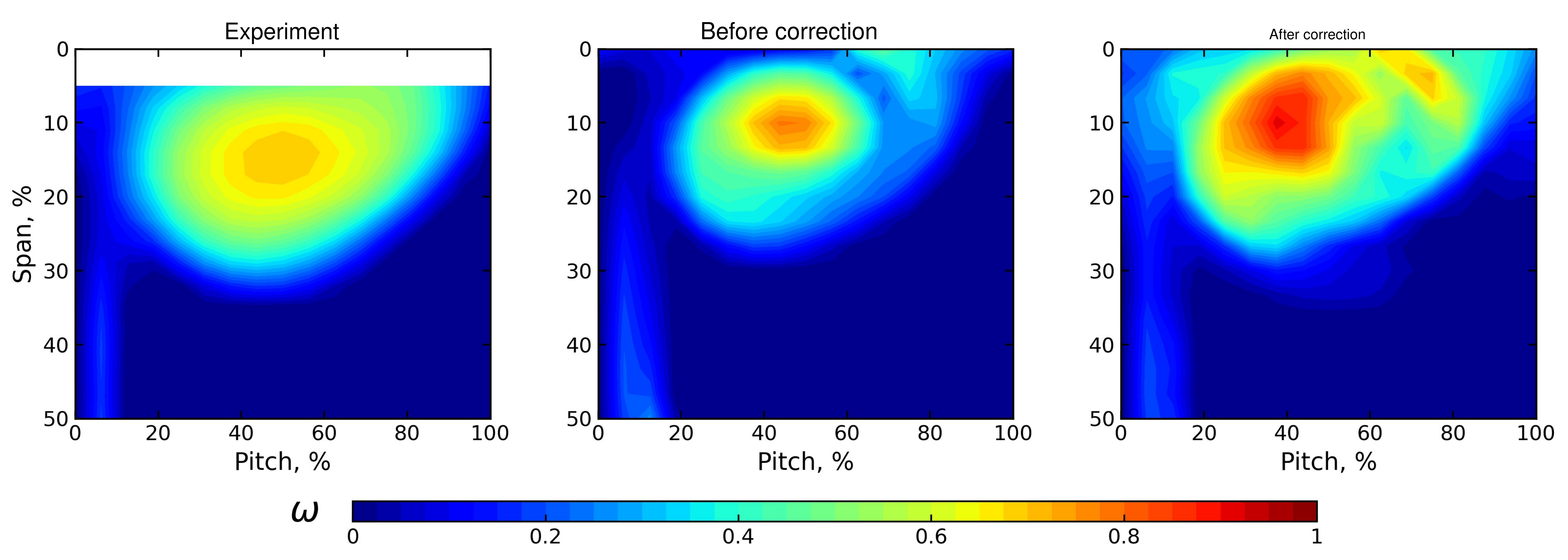}
\caption{i = 6\ensuremath{^\circ}, s = 5\% of blade span, Ma = 0.3, n = 0 rpm}
\end{subfigure}
\par\smallskip
\begin{subfigure}{0.88\textwidth}
\centering
\includegraphics[width=\linewidth,height=0.19\textheight,keepaspectratio]{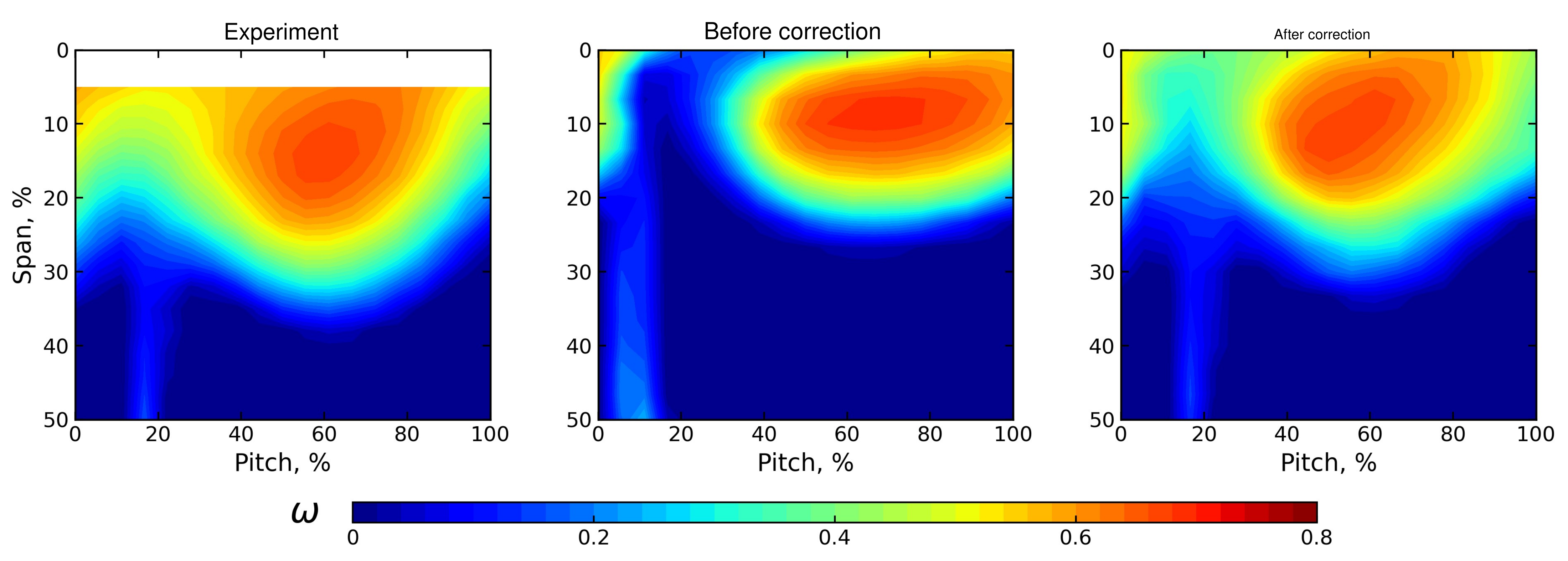}
\caption{i = 6\ensuremath{^\circ}, s = 7.14\% of blade span, Ma = 0.3, n = 900 rpm}
\end{subfigure}
\par\smallskip
\caption{Experimental observations and CFD-based predictions before and after correction at four operating conditions.}
\label{fig:9}
\end{figure}

Figure~\ref{fig:9} compares experimental observations with the CFD-based fields before and after correction for several operating conditions.

Although the uncorrected VAE reconstructions capture the principal high-loss regions, their peak locations, spatial extents, and gradients differ noticeably from the experimental observations. After latent-space correction, the red and orange high-loss regions generally move closer to the measurements in both position and coverage while maintaining smooth spatial transitions. All four illustrated conditions show improvement. These results indicate that the adapter adjusts local high-loss structures under the measurement constraints rather than simply applying a uniform scaling to the total pressure loss coefficient.

\section{Conclusions}

This study addresses the correction of CFD total pressure loss fields for open tip clearance flow in a compressor cascade using experimental observations. CFD data can be generated in batches at high spatial resolution, but predictions differ from measurements. Experimental data more directly represent the actual flow response, yet acquisition cost and spatial access limit the number of paired conditions and confine measurements to discrete locations. To accommodate these heterogeneous data sources, a non-intrusive correction method is developed using a VAE with latent-space adaptation and an explicit observation operator. The main conclusions are as follows.

(1) The Stage 1 VAE provides an effective low-dimensional representation of total pressure loss fields. For 25 CFD samples excluded from training, the reconstructions preserve the principal locations and extents of high-loss regions and their overall spatial gradients. MAE and RMSE are relatively tightly distributed across samples. Errors occur mainly near loss peaks and the edges of high-loss regions, indicating that the VAE adequately represents the dominant field structures.

(2) The explicit sparse observation operator establishes spatial correspondence between high-resolution CFD predictions and discrete experimental observations. By mapping the corrected field to the actual measurement locations and pitchwise windows, the model learns the CFD--experiment discrepancy directly in observation space and backpropagates the measurement error to the latent-space adapter. This avoids interpolating sparse measurements into artificial high-resolution labels and the additional errors that such full-field label construction could introduce.

(3) Latent-space adaptation substantially reduces observation-space prediction errors for unseen paired conditions. In the current 12-fold results, the mean foldwise improvement exceeds 62.00\% for multiple metrics. The similar improvements in MAE and RMSE indicate reductions in both the overall measurement discrepancy and larger local errors.

The method thus enables CFD correction using sparse experimental data. The VAE prior provides high-resolution structural constraints for open tip clearance loss fields, while experimental observations constrain corrections through adaptation of the latent representation. Their combination supports non-intrusive field correction with a small number of paired samples.

\section*{Funding}

This work was supported by the Aeronautical Science Foundation (No. 2024L039057003).

\end{document}